\documentclass[final, runningheads]{llncs}
\usepackage[T1]{fontenc}
\usepackage{graphicx}
\usepackage{amsmath}
\newcommand{\itadata}{\footnotesize \textsl{ITADATA2024: The 3$^{\text{rd}}$ Italian Conference on Big Data and Data Science}}
\usepackage{fancyhdr}
\fancypagestyle{empty}{\fancyhf{}\fancyhead[C]{\itadata}
  \fancyhead[R]{\includegraphics[width=.05\textwidth]{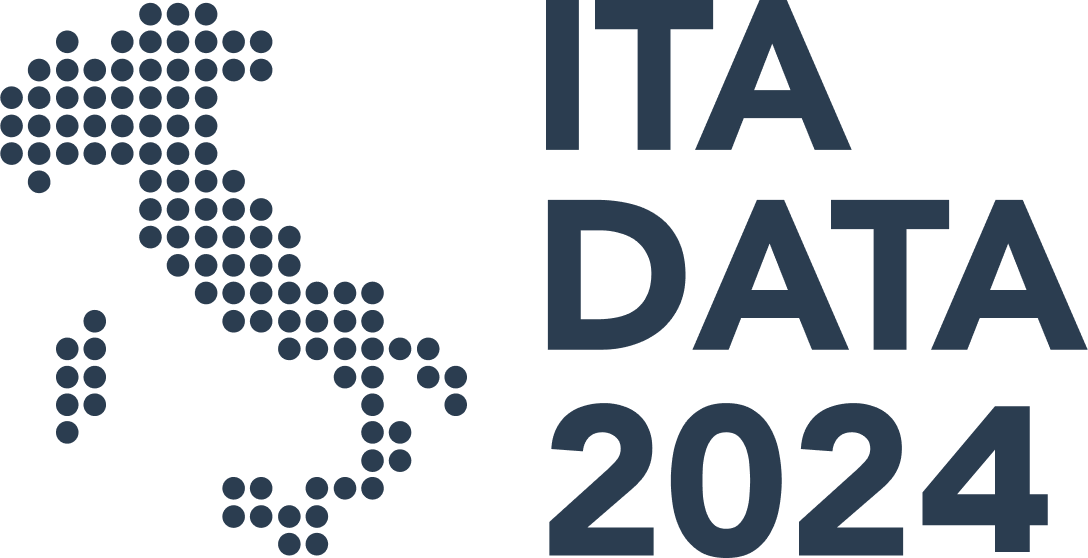}}}

\makeatletter
\begin{document}
\title{A Methodology to extract Geo-Referenced Standard Routes from AIS Data}
\author{Michela Corvino\inst{1} \and
Filippo Daffinà\inst{2} \and
Chiara Francalanci\inst{3} \and
Paolo Giacomazzi\inst{3} \and
Martina Magliani\inst{4} \and
Paolo Ravanelli\inst{4} \and
Torbjorn Stahl\inst{2}}
\authorrunning{F. Author et al.}
\institute{\
ESA, Via Galileo Galilei, 00044 Frascati, Italy \email{michela.corvino@esa.int}\and
e-Geos, Via Tiburtina, 965, 00156 Rome, Italy \email{\{filippo.daffina, torbjorn.stahl\}@e-geos.it}\and
Politecnico di Milano, P.zza Leonardo da Vinci, 32, 20133 Milan, Italy \email{\{chiara.francalanci, paolo.giacomazzi\}@polimi.it}\and
Cherrydata, Via Abano, 9, 20131 Milan, Italy \email{\{martina.magliani, paolo.ravanelli\}@cheerry-data.com} }
\maketitle              \begin{abstract}
    Maritime AIS (Automatic Identification Systems) data serve as a valuable resource for studying vessel behavior. This study proposes a methodology to analyze route between maritime points of interest and extract geo-referenced standard routes, as maritime patterns of life, from raw AIS data.
The underlying assumption is that ships adhere to consistent patterns when travelling in certain maritime areas due to geographical, environmental, or economic factors. Deviations from these patterns may be attributed to weather conditions, seasonality, or illicit activities. This enables maritime surveillance authorities to analyze the navigational behavior between ports, providing insights on vessel route patterns, possibly categorized by vessel characteristics (type, flag, or size).
Our methodological process begins by segmenting AIS data into distinct routes using a finite state machine (FSM), which describes routes as segments connecting pairs of points of interest. The extracted segments are aggregated based on their departure and destination ports and then modelled using iterative density-based clustering to connect these ports. The clustering parameters are assigned manually to sample and then extended to the entire dataset using linear regression. Overall, the approach proposed in this paper is unsupervised and does not require any ground truth to be trained.
The approach has been tested on data on the on a six-year AIS dataset covering the Arctic region and the Europe, Middle East, North Africa areas. The total size of our dataset is 1.15 Tbytes. The approach has proved effective in extracting standard routes, with less than 5\% outliers, mostly due to routes with either their departure or their destination port not included in the test areas.

\keywords{big data \and AIS data \and maritime route extraction \and maritime patterns of life \and maritime domain awareness \and density-clustering \and unsupervised learning.}
\end{abstract}
\section{Introduction}
The paper aims to develop a methodology for extracting standard vessel routes from AIS data, focusing on vessels departing from and arriving at points of interest (POIs), such as ports, offshore slots, or oil platforms. Maritime analytics, including standard route extraction, is gaining traction due to the significant role of sea trade in global commerce. However, the lack of ground truth and of a clear operational definition of standard routes in the literature has led us to the development of a new concept of geo-referenced standard route. This concept defines standard routes as trajectories connecting POIs, thus facilitating the extraction of meaningful information from AIS data.

The methodology employs an automated step-by-step approach to segment raw AIS data into distinct routes based on maritime POIs. This segmentation method effectively captures ship behaviors that are not clear from raw AIS data alone. For each vessel, segments are grouped based on departure and destination POIs. For each group, one or multiple standard routes are extracted using an iterative density-based clustering approach that can identify all distinct paths commonly followed by ships between these POIs. This incremental learning approach dynamically does not require any ground truth, allowing standard routes to be extracted using only AIS data without prior information.

This methodology can contribute to ship position prediction for Maritime Situational Awareness (MSA), raising the need for longer time intervals beyond existing monitoring capabilities. Leveraging standard vessel patterns, this research could help extend the time horizon of the prediction of ship positions beyond current 3–6-hour standards, facilitating port-to-port trajectory reconstruction. In turn, this could provide benefits ranging from better search and rescue operations, to automated alerting mechanisms showing anomalous ship behavior, considering seasonality as well as the specific features of ships and geographical areas. This methodology could also offer an effective solution for MSA authorities, enabling the comprehensive analysis of vessel behavioral patterns to understand common trends.

The presentation is organized as follows. Section 2 reviews the state of the art, with a focus on previous research on AIS data. Section 3 presents our approach to the extraction of standard routes, providing a short description of the system that has been designed. Section 4 summarizes empirical results and discusses the main findings. Conclusions are finally drawn in Section 5.

\section{State of the Art}
The existing literature lacks a precise definition of standard route, but various methods aim to extract motion patterns representing typical vessel behaviors within specific timeframes and areas of interest. These patterns, based on real-time positional data including vessel longitude, latitude, and speed, are essential for accurately predicting vessel positions. Prior research outlines a systematic approach to address this challenge, involving several key steps:
\begin{enumerate}
    \item Data Preprocessing: AIS data undergoes preprocessing to ensure data quality.
    \item Clustering: Historical motion data are clustered using techniques such as TREAD [8] to group data into motion patterns.
    \item Vessel Assignment: Vessels are assigned to clusters based on their behavior and comparison with motion patterns.
    \item Trajectory Prediction: Using the assigned cluster’s route pattern, vessel positions are predicted, with accompanying anomaly detection and alerting services.
\end{enumerate}

The literature presents various approaches for extracting motion patterns from AIS data, facilitating vessel position prediction and enhancing maritime security efforts. One approach, outlined in [6], employs a statistical model to learn motion patterns without prior knowledge of traffic scenes. This method includes grid-based and vectorial representations, with the latter connecting waypoints and turning points to define trajectories. However, challenges arise in detecting turning points along complex routes. To address this, Pallotta et al. introduce the TREAD algorithm [8], using density-based clustering to detect waypoints and extract similar trajectories. Another method, presented in [9], applies K-means clustering to geographical regions to identify clusters of data with similar vessel speed and heading direction, further analyzed using a Hidden Markov Model [9]. Hierarchical clustering is employed in [4] to group trajectories based on spatial and temporal information. Notable examples of motion extraction techniques include the Ornstein-Uhlenbeck Processes [2], Bayesian prediction approaches [7], and Gaussian mixture models [1].

Our approach distinguishes itself from the existing literature in several key aspects. First, we employ an unsupervised machine learning approach, specifically a trajectory-based method, to extract complex port-to-port trajectories. This approach eliminates the need for prior knowledge or labeled datasets, unlike other methods that rely on supervised learning or point-based trajectory extraction techniques. Additionally, it does not require real-time information and allows for a comprehensive analysis of historical data, with a potentially positive impact on the long-term prediction of vessel trajectories. Furthermore, our system incorporates robust data pre-processing techniques to ensure cleaner, higher-quality data, leading to more accurate ground truth for machine learning algorithms. 

By dividing the AIS messages by vessel types before training, we capture motion patterns specific to each vessel type, preventing the mislabeling of normal behaviors as anomalous. Moreover, we have integrated a geographical component with port coordinates, facilitating port-to-port trajectory reconstruction, unlike the position-based approach followed by [1–3, 5, 7]. Notably, this feature is exclusively found in [9], where it is primarily used for port traffic monitoring purposes. In contrast to existing methods that may resample trajectories or remove outliers before training, our approach retains outliers until identified by the algorithm, preserving crucial data. Overall, our approach addresses several gaps in the existing literature, to offer a comprehensive solution for standard route extraction and vessel position prediction in maritime contexts.

\section{State of the Art}
This section presents our methodology, which defines our concept of standard route as the extraction of a unique route from the aggregation of port-to-port segments. From an initial unique segment made up by all the AIS messages transmitted from the same MMSI (Maritime Mobile Service Identity) we derive ports and segments, through a finite state machine (FSM) dividing each time series into segments whenever a ship arrives in a port or if it stops transmitting for a significant amount of time. A key innovation in our approach lies in the idea that meaningful ports must represent the end and the start of segments. Consequently, the accuracy of the port database is crucial in our methodology.

\begin{figure}
\centering
\includegraphics[width=0.4\textwidth]{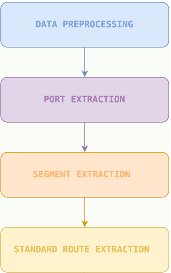}
\caption{Methodology for naval route extraction: data processing pipeline.} \label{fig1}
\end{figure}

The methodology pipeline is structured into distinct steps designed to implement the route extraction process, as depicted in Figure 1. The process begins with data pre-processing, followed by port extraction, segment extraction, and standard route extraction. Each step is designed to cope with the complexity of maritime route analysis, while maintaining analytical precision. The next sub-section presents the four methodological steps shown in Figure 1.

\subsection{Data pre-processing}
The first phase in our methodology pipeline is data pre-processing, which comprises three key steps: feature selection, identification of data quality issues, and vessel classification. Feature selection involves a careful choice of the essential dynamic and static features from AIS data that are conceptually necessary to achieve our route extraction goal. In our approach, these features include MMSI, timestamp, latitude, longitude, vessel type, flag country, destination, and navigational status. By focusing on this subset of features, we ensure that our subsequent analyses are based on the most relevant information in AIS messages, thus reducing unnecessary complexity. 

Identifying data quality issues is another critical aspect of the pre-processing phase. This activity involves recognizing and addressing various issues that may affect the reliability and accuracy of AIS data. Common issues include inconsistent message reception times, inaccurate geographical coordinates, unreliable navigational status fields, and inconsistencies in redundant information. By addressing these issues upfront, we can minimize errors and improve the quality of our dataset. 

Vessel classification is the final step in the preprocessing phase, where we categorize vessels based on specific characteristics. This classification allows us to identify common behavioral patterns within each category, leading to more coherent aggregate data. As an example, fishing boats follow navigational patterns that go through the best fishing spots, while cargo ships are more likely to follow either the shortest or the safest path between POIs.

\subsection{Port extraction}
As noted, an important step in our methodology is to build the dataset of the maritime geographical points of interest. In the port extraction phase, we have first analyzed open-source ports databases, namely Open Street Map (OSM) and World Port Index (WPI). We have noted how they lack completeness, as in our AIS datasets ships would declare to be in a port in locations that were not classified as ports in these widely used port databases. Therefore, we have developed a custom solution. This solution leverages vessel movements in combination with significant changes in a ship’s heading to identify potential port locations. With this approach, we can classify as ports numerous on shore locations that we could not extract from official port databases as well as various offshore slots. Specifically, our methodology entails a systematic examination of AIS messages to analyze the vessel’s position, speed and heading direction over time. We calculate the bearing between subsequent geographical coordinates when the algorithm detects an average speed window under a specific threshold (MIN$\_$SPEED$\_$DEPARTURE), to identify changes in heading that may be an indication of a mooring manouver.

\begin{figure}
\centering
\includegraphics[width=\textwidth]{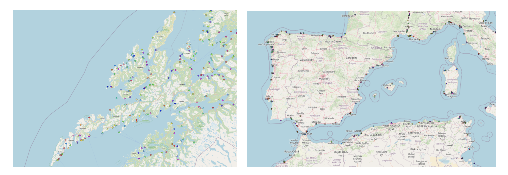}
\caption{Visualization of extracted ports (example).} \label{fig2}
\end{figure}

By identifying heading changes, our algorithm can detect ports not only in harbor areas but also in open water regions. This is attributed, in part, to the fact that many vessels do not dock directly at ports but have predefined offshore slots, as depicted in Figure 3. We have used the open-source port databases to label as many as possible of our port locations. We have been able to label roughly 50\% of all ports extracted with our approach.

\begin{figure}
\centering
\includegraphics[width=\textwidth]{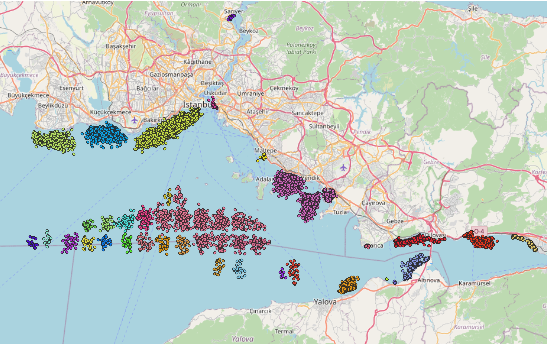}
\caption{Representation of offshore slots.} \label{fig3}
\end{figure}

\subsection{Segment extraction}
In this phase, the objective is to extract segments delimited by ports retrieved in the previous phase and then reduce them by destination to decrease the percentage of partial routes caused by GPS errors, anomalous stops in transmitting data or other data-related inconsistencies. Data are modeled according to a finite state machine (FSM, see Figure 4). The FSM is designed to analyze each vessel movement and identify port-to-port segments through a series of states and transitions. These states, including initialization, departure, sailing, stationary, arrived, and lost, are determined based on various criteria such as vessel speed, location, and data availability. Each state transition is triggered by specific conditions related to these criteria, guiding the algorithm through the segmentation process. In detail, the states in the FSM are: 

\begin{itemize}
    \item INIT: Initialization state.
    \item DEPARTURE: Vessel is departing from a port.
    \item SAILING: Vessel is in transit.
    \item STATIONARY: Vessel is stationary.
    \item ARRIVED: Vessel has arrived at a port.
    \item LOST: Vessel information is lost or inconsistent.
\end{itemize}

Each MMSI is assessed according to the FSM shown in Figure 4.

\begin{figure}
\centering
\includegraphics[width=\textwidth]{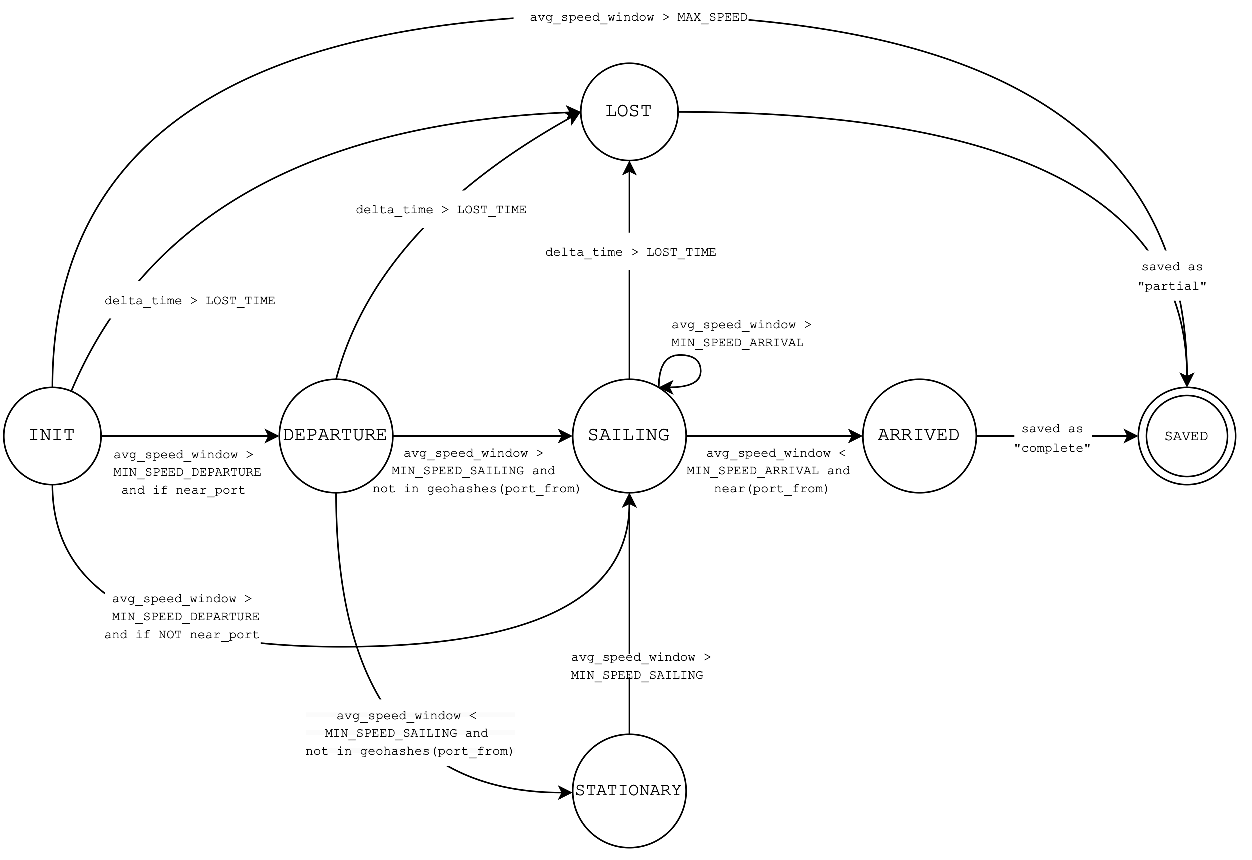}
\caption{Finite state machine used to extract segments from AIS messages.} \label{fig4}
\end{figure}

The output of this segmentation is given as input to a segment reduction by destination step focused on refining the extracted segments to create more accurate and complete routes. This algorithm works by merging consecutive segments with the same destination, without arrival or departure ports, effectively reducing redundancy and improving the overall quality of the route dataset (see example in Figure 5). Overall, the output of this phase is a comprehensive route dataset for each distinct MMSI.

\begin{figure}
\centering
\includegraphics[width=\textwidth]{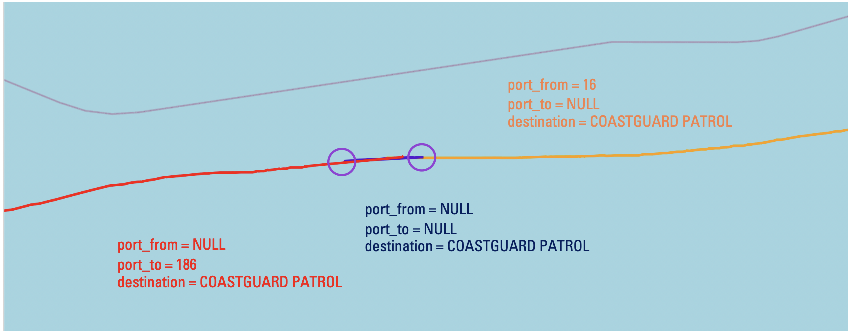}
\caption{Three different segments representing a route from Norway, Sortland to Coastguard Patrol, Iceland.} \label{fig5}
\end{figure}

\subsection{Standard route extraction}
The input of this phase are the route datasets for each MMSI and the related port coordinates. These port coordinates are specific to each MMSI so we need to create a comprehensive dataset of ports by clustering all the ports of all the MMSIs, using a density-based technique, to accurately extract clusters of points that represent different locations of ports where the vessels are usually docked. The centroid is taken as the reference point for each cluster. The OSM dataset of ports is subsequently integrated to further enhance accuracy. This step helps in recognizing different departure points within the same port. After creating our dataset, we update each MMSI route with the new cluster centroid coordinates. In this way, we can aggregate all the MMSI routes based on departure and destination ports, having as output an aggregated dataset with all the routes extracted with the same departure and destination port. Based on this output, we can finally derive meaningful standard routes. The approach used aims to generate standard routes with an iterative DBSCAN-based approach to cluster points and build a standard route. The steps followed to obtain a standard route are as follows:

\begin{enumerate}
    \item The standard route begins with the coordinates of the departure port.
    \item All points within a specified radius r are selected. The searching process continues until enough points are selected, ensuring that all points of the routes are visited.
    \item A DBSCAN algorithm is applied to find possibly multiple clusters of points.
    \item For each cluster identified by the DBSCAN algorithm, the barycenter is computed.
    \item The algorithm iteratively extends the standard route with the computed barycenters.
    \item If more than one cluster is identified by the DBSCAN, we create as many standard routes as the number of clusters identified, each with the barycenter of one of the identified clusters.
    \item These steps are reiterated until no additional points are identified.
\end{enumerate}

Our methodology also includes the following key components:

\begin{itemize}
    \item Handling edge cases: The methodology incorporates mechanisms to handle edge cases such as excessively short routes or no additional points within an expanded radius. Routes are marked as completed only if they stop near the destination port. 
    \item Cluster labeling and route splitting: In scenarios with multiple clusters identified by the iterative density clustering algorithm, indicating potential route variations, the methodology facilitates the segmentation of standard routes into distinct ones, each associated with a unique label. 
    \item Input parameters determination: Determining optimal values for epsilon ($\epsilon$) and the minimum number of samples for the DBSCAN algorithm, as well as the appropriate research radius ($r$) is crucial. This involves a one-time manual evaluation of parameters for a small subset of aggregated routes, considering factors such as spatial and temporal characteristics, number of routes, median duration, and average distance traveled. 
    \item Machine learning for parameter determination: To scale the process and determine the DBSCAN parameters for all aggregate routes, a semi-supervised linear regression model is trained using manually computed target values and selected characteristics of aggregate routes as predictive variables. 
\end{itemize}

Meaningful characteristics such as spatial and temporal sampling, number of points, and distance traveled per route are computed for each set of aggregate routes, informing the manual evaluation of input parameters. The methodology is divided into distinct sub-parts, including spatial and temporal sampling computation, manual evaluation of parameters, and machine learning-based determination of unknown parameters. This will help the application of the methodology in real scenarios and its integration in platforms for MSA authorities.

\section{Testing}
During the testing and validation phase, our methodology has been evaluated to ensure its effectiveness and reliability. The architecture of the tool, depicted in Figure 6, encompasses several key components, each playing a specific role in processing AIS data and extracting meaningful standard routes.

The methodology validation process involves assessing each component of the tool architecture individually to ensure its effectiveness in achieving the desired outcomes on a six-years AIS dataset covering the Arctic region and The Europe, Middle East, North Africa area.  

\begin{figure}
\centering
\includegraphics[width=\textwidth]{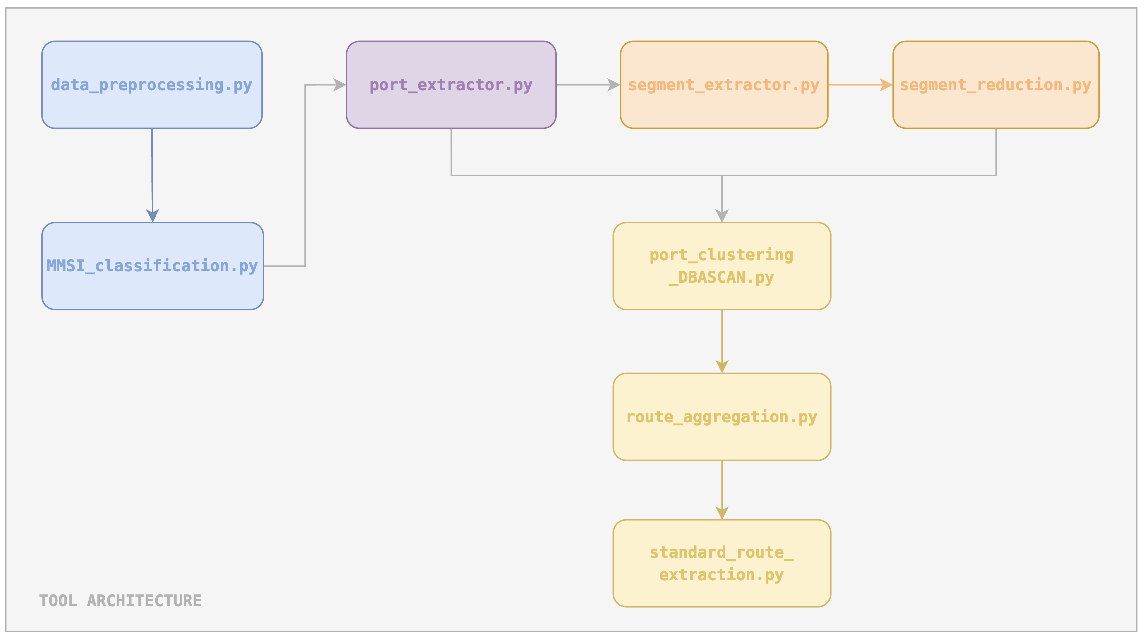}
\caption{Architecture of the tool.} \label{fig6}
\end{figure}

During the data preprocessing phase, our evaluation focuses on how effectively the dataset size is reduced by retaining only relevant features, and how issues related to missing or erroneous information are addressed. The data preprocessing phase efficiently transformed the Arctic and EMENA datasets, originally totaling 150 Gbyte and 1 Tbyte respectively, into structured formats suitable for analysis, resulting in size reductions of approximately 77.65\% and 80.65\%. These processed datasets, now 33.53 Gbyte and 193.5 Gbyte in size respectively, served as optimized inputs for subsequent phases of the methodology.

In the port extraction phase, we assess the quality of the final port dataset by visualizing and comparing it with those of open port datasets (i.e. OSM, WPI) and based on the algorithm’s ability to correctly delineate segments with extracted ports. Since the complete routes represent the 95.75\% of the total, the port dataset proves its effectiveness and accuracy. The remaining 4.25\% are partial routes without a departure or an arrival port, which could be caused by missing data or, most often, by the limited boundaries of the considered AOIs.

The segment extraction phase is evaluated through visualization by assessing the effectiveness of route aggregation into one cluster with common departure and arrival ports. The results are categorized for each vessel type and prove a high level of precision and accuracy for all vessel types in determining the departure/arrival of a vessel from/to a specific port area.

\begin{figure}
\centering
\includegraphics[width=0.6\textwidth]{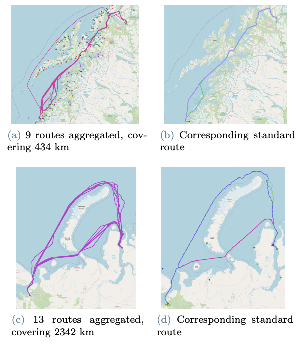}
\caption{Sample results.} \label{fig7}
\end{figure}

The assessment of the standard route extraction phase is based on the accuracy of routes reaching their destination ports. In an end-to-end qualitative evaluation of the methodology for extracting standard routes, a sub-sample of 300 Cargo type routes was analyzed out of 12104 that were created by the standard route extraction algorithm. It was found that 92\% of the routes reached the destination port correctly, while 89\% correctly split the standard route following the expected patterns. The remaining 11\% of standard routes were typically composed of aggregate data with a low number of routes, resulting in a small number of points, which increased the likelihood of failure in detecting clear clusters. However, most routes accurately represented the expected navigation patterns. In terms of quantitative evaluation, it was determined that 95\% of the standard routes reached the destination port. Given that the subset manually analyzed showed 89\% high-accuracy results, it is expected that the majority of the 95\% of standard routes correctly reached their destination ports.

Computational time issues emerged during the process but were effectively resolved by computing standard routes in parallel on different cores of the same server. This optimization significantly reduced computation time, enabling a more efficient use of resources, and expediting the standard route extraction process for the entire dataset. Overall, these results highlight the efficiency and effectiveness of the methodology employed for creating standard routes, with areas for improvement identified for future enhancements, as discussed in the next section.

\section{Concluding Remarks}
In this study, we propose an innovative approach aimed at enhancing maritime security through route aggregation, using AIS data across diverse Areas of Interest (AOIs) spanning extensive geographical regions. Our objective was to understand and study ships’ behavioral patterns derived from AIS data, evaluating our concept of standard route using a six-year AIS dataset encompassing the Arctic region and the Europe, Middle East, and North Africa area. Our approach for route aggregation and standard route extraction, leveraging ports as segment delimiters, has been implemented with the Tool Architecture presented in Figure 6 and then tested on our AOIs.

Despite encountering significant challenges related to big data and data quality, our overall approach has proved effective. Our method’s advantage lies in its automated step-by-step approach, providing a visual output at each stage, facilitating parameter fine-tuning, and offering a comprehensive understanding of the available data. Additionally, by focusing on specific vessel types, our method facilitates the detection of different behaviors for each vessel type, enabling segmentation based on desired features such as flag type or vessel size. To validate the end-to-end methodology, we conducted a manual visualization of the standard route extraction output, analyzing selected examples representing the most challenging cases. These examples allowed us to gauge the algorithm’s performance in different scenarios, particularly highlighting challenges in identifying consistent patterns for highly irregular or short trajectories, such as those typical of fishing vessels.

First of all, future work will address minor issues in the production of standard routes, with reference to 1) improving the adherence of standard routes to river courses, 2) making sure that standard routes never pass over land, and 3) addressing the borders of the AOI not to discard routes with no destination port (that is exiting the AOI). Future research could leverage the standard routes dataset to train machine learning models for predicting vessel trajectories in cases where AIS data transmission stops. Predicting a vessel’s likely destination within a 12–24-hour timeframe, based on its departure port and previous trajectory, holds significant potential for maritime surveillance efforts. This predictive capability, coupled with Earth Observation Synthetic Aperture Radar (EO SAR) systems, could be instrumental in monitoring and combating illegal commercial activities, piracy, and other illicit operations, enhancing maritime security on a global scale.

\begin{credits}
\subsubsection{\ackname} This research has been partly supported by the European Space Agency within the EO4SECURITY project (INNOVATIVE SAR PROCESSING METHODOLOGIES FOR SECURITY APPLICATIONS) in the framework of the European Space Agency Contract No. 4000142273/23/I-DT.
\end{credits}

\end{document}